# 1-degree-of-freedom robotic gripper with infinite self-twist function

Toshihiro Nishimura[1], Member, *IEEE*, Yosuke Suzuki[1], *Member, IEEE*, Tokuo Tsuji[1], *Member, IEEE*, and Tetsuyou Watanabe[1], *Member, IEEE*

*Abstract*— **This study proposed a novel robotic gripper that can achieve grasping and infinite wrist twisting motions using a single actuator. The gripper is equipped with a differential gear mechanism that allows switching between the grasping and twisting motions according to the magnitude of the tip force applied to the finger. The grasping motion is activated when the tip force is below a set value, and the wrist twisting motion is activated when the tip force exceeds this value. "Twist grasping," a special grasping mode that allows the wrapping of a flexible thin object around the fingers of the gripper, can be achieved by the twisting motion. Twist grasping is effective for handling objects with flexible thin parts, such as laminated packaging pouches, that are difficult to grasp using conventional antipodal grasping. In this study, the gripper design is presented, and twist grasping is analyzed. The gripper performance is experimentally validated.**

## I. INTRODUCTION

THIS study proposed a novel underactuated robotic gripper that can achieve grasping and infinite wrist twisting motions using a single actuator. Fig. 1(a) illustrates the gripper motion when the motor is rotated unidirectionally. With the use of a differential gear mechanism, the motion of the gripper can be switched between grasping (closing/opening of the gripper) and twisting, depending on the magnitude of the load applied to the fingertip. Using this mechanism, the gripper achieves conventional antipodal grasping and a special grasping mode (twist grasping) through wrist twisting. During twist grasping, an object with a flexible thin part is wrapped around the finger by twisting the gripper, as shown in Fig. 1(b). This grasping mode enables stable grasping and a large payload for grasping an object with flexible thin parts such as laminated packaging pouches. In environments such as supermarkets where items used in daily life are handled, laminated packaging pouches are among the most difficult objects to grasp. Generally, pouch packages include flowable objects such as granular materials and liquids. This flowability deforms the pouch, and its center of gravity easily changes during grasping. Additionally, certain pouches are heavy. In supermarkets, pouches are generally densely placed. Typically, the accessible area of the target pouch is the top thin part, unless complex operations such as moving obstructing pouches out of the way to approach the target pouch are performed. These conditions reduce the stability of conventional antipodal grasping for pouches in practical situations. To overcome this issue, we propose twist grasping, which is similar to the working of a human hand. We develop a robotic gripper that realizes both twist and antipodal grasping using a single actuator.

Underactuation has the benefits of a simple control scheme and lightweight compact design. Wrist rotation is effective for handling objects, e.g., aligning object postures and tightening screws. Considering that few commercial manipulators provide an unlimited rotational motion range in the most distal joint, an end effector with an infinite wrist rotation function can expand the operation range of the manipulators.

Thus, we were motivated to develop a single-actuator gripper with a wrist rotation function. Our design goals are as follows: 1) to realize the grasping (closing/opening motion of the fingers) and twisting motions of the wrist using a single actuator; 2) to grasp a heavy object (>10 kg) with a flexible thin part; and 3) to achieve a lightweight design (<0.5 kg).

### A. Related Work

Several underactuated robotic grippers that can switch between functions (including grasping modes) using a single actuator have been proposed [1]. In previous studies, the contact between fingers and the object has been utilized to switch between functions. Kakogawa et al. designed a robotic gripper

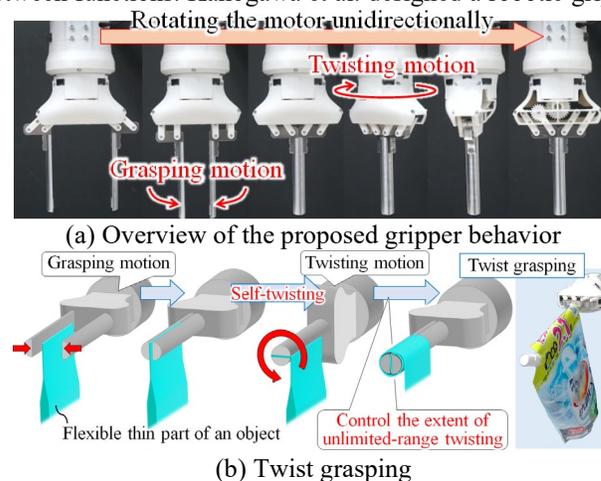

(a) Overview of the proposed gripper behavior

(b) Twist grasping

**Fig. 1.** Proposed gripper motion and twist grasping

Manuscript received: February 24, 2022; Revised: May 25, 2022; Accepted June 13, 2022. This letter was recommended for publication by Associate Editor and Editor Liu Hong upon evaluation of the reviewers' comments.

This work was supported by JSPS KAKENHI Grant Number 22K14220 *(Corresponding author: T. Nishimura).*

[1]T. Nishimura, Y. Suzuki, T. Tsuji, and T. Watanabe are with the Faculty of Frontier Engineering, Institute of Science and Engineering, Kanazawa University, Kakuma-machi, Kanazawa city, Ishikawa, 9201192 Japan (e-mail: tnishimura@se.kanazawa-u.ac.jp, te-watanabe@ieee.org).

Digital Object Identifier (DOI): see top of this page.

that can switch between grasping motion and object-pull-in functions through contact with an object by using a differential gear mechanism [2]. Ko proposed a tendon-driven robotic gripper with a pull-in function; the mode of the gripper could be changed by contact with the target object [3]. Similarly, we proposed single-actuator robotic grippers that can switch between multiple grasping modes by utilizing contact with supporting surfaces such as a table [4][5]. A typical mechanism that can passively change grasping postures is the underactuated finger mechanism with a self-adaptive function that passively changes the finger posture upon contact with an object [6]–[14]. A change in the grasping posture corresponds to switching between functions. Several single-actuator robotic grippers that can switch functions without contacting the surroundings have been proposed [15]–[18]. However, they require complex mechanisms/control strategies and are difficult to apply to mechanisms that rotate the wrist. In [11]–[13], differential gear mechanisms were installed to switch between functions, as in this study. However, these studies did not focus on wrist rotation.

Several robotic wrist mechanisms have been proposed [19]. Certain wrist mechanisms realize twisting motions to expand the operating range of a manipulator [20]–[23]. However, in these mechanisms, the number of actuators used for wrist or grasping motions is not minimized. A robotic wrist that achieves the twisting motion for a peg-in-hole without installing an actuator has been proposed [24]. However, a manipulator is needed to activate the twisting motion, and the twist angle that could be generated from a single activation is limited to a finite value.

A few studies have proposed robotic grippers with wrist mechanisms that can achieve infinite wrist twisting [25][26]. Hughes et al. proposed a "rotary gripper" that achieves grasping motion and infinite wrist twisting using two motors [25]. Additionally, a parallel-jaw gripper with a wrist rotation mechanism similar to that of [25] is commercially available [26], indicating an increased demand for the wrist twisting function.

Although several studies have developed grippers with mechanisms that passively switch between functions, no attempt has been made to realize a robotic gripper with a wrist mechanism that rotates infinitely by using a single actuator.

## II. TWIST GRASPING AND GRIPPER DESIGN

### A. Twist Grasping

Twist grasping is a special grasping mode realized using the proposed gripper. The procedure for grasping a flexible thin object via twist grasping is illustrated in Fig. 1(b). The object is initially grasped via antipodal grasping and then wrapped around the fingers of the gripper by twisting the wrist. The switch from grasping to twisting is triggered by an increased load on the fingertip due to the contact between the finger and the object. Switching is achieved using a single actuator; thus, the twisting function is considered a self-twist function. The proposed gripper can hold heavy objects via twist grasping, because the payload due to twist grasping increases with the number of wraps (see Section III for details). As the twisting range is infinite, the payload due to twist grasping can be infinitely increased as long as the gripper does not break.

### B. Functional Requirements

The functional requirements of the proposed robotic gripper for achieving antipodal and twist grasping with a single actuator are as follows: 1) a mechanism to achieve finger movements and infinite wrist rotations using a single actuator; 2) the weight of a target object with flexible thin parts should be >10 kg; and 3) a lightweight design (<0.5 kg including the motor).

The gripper design and control methodology for switching between the grasping modes are presented below.

### C. Principal Mechanism

Figs. 2 and 3 show the three-dimensional computer-aided design (3D-CAD) model and driving mechanism for switching the finger and wrist motions of the gripper, respectively. The gripper consists of rotation and fixing units (Fig. 3). The rotation unit includes gear 2 (yellow parts in Fig. 2), gear 3 (blue parts), gear 4 (red parts), a gearbox (light pink part), fingers (light blue part), and finger links (gray parts). The fixing unit includes a fixed base (translucent gray part), a motor (black part), gear 1 (green part), and a preloader (orange and purple parts). The rotation unit is mounted on the fixing unit via a bearing. Gear 2 is a pair of double gears, each consisting of a bevel gear meshing with gear 1 and a spur gear meshing with gear 3. The motor torque is transmitted to gear 2 via gear 1. Gear 4 is unified with a link to move the finger (red). To generate the finger movement, a four-bar linkage comprising a gear 4 link and finger links is used. There are two sets of gears 2, 3, and 4, and finger links are symmetrically arranged with respect to the rotational axis of gear 1. When the pair of gears in gear 2 rotates in opposite directions, opening and closing motions of the fingers are generated. A grasping motion is generated if the gearbox is fixed. If the gearbox is free, the rotation unit rotates, and a twisting motion occurs. A preloader is installed to achieve the desired mechanism, that is, switching between grasping and twisting motions according to the load due to the contact between the fingertip and the object. The preloader includes an indenter (orange part in Fig. 2), a fixer (purple part), and a set screw. The indenter and fixer have male and female threads, respectively. The fixer is mounted on a fixed base. A preload is applied to the gearbox by screwing the indenter to prevent the rotation unit from rotating. The preload can be tuned by changing the screwing depth of the indenter and can be kept constant by holding the indenter at its screwed position using the set screw. As shown in Fig. 3, when the fingertip does not make contact with an object, the rotation unit is not rotated owing to the preload; then, the input torque from the motor is transmitted to gear 4, generating the finger motion. If the fingers make contact with an object and the force applied to the fingertips by the object exceeds the value needed to release the motion constraint on the gearbox due to the preload, the finger movement stops. Correspondingly, the axial rotation of gear 2 stops. Subsequently, the input torque is transmitted to the gearbox via gear 2, and the rotation unit is rotated. The tip force is maintained when the rotation unit is rotated or twisted.

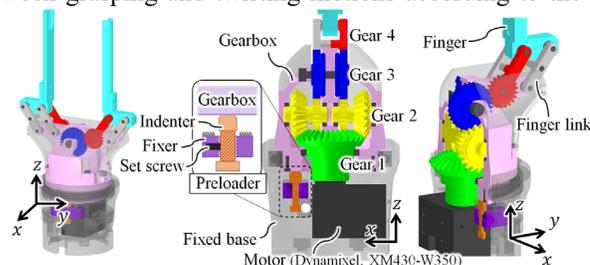

**Fig. 2.** 3D-CAD model of the developed gripper

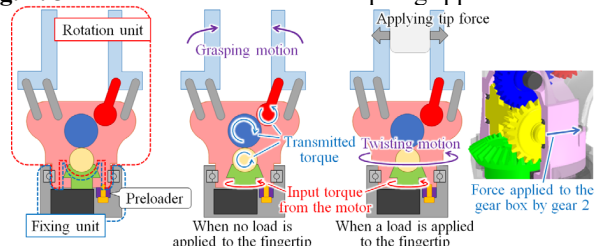

**Fig. 3.** Schematic of the driving mechanism

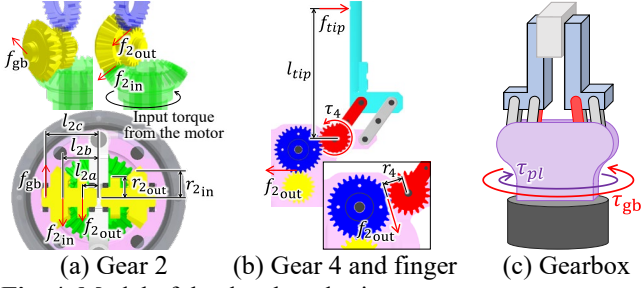

(a) Gear 2  (b) Gear 4 and finger  (c) Gearbox

**Fig. 4.** Model of the developed gripper

*D. Statics*

In this section, the relationship between the force on the fingertip and the motor torque is derived, along with the condition for the switching between the two grasping modes: grasping and twisting. This condition is used to estimate the state of the gripper and control mode switching using the motor torque, which can be derived from the motor current. The nomenclature used is shown in Fig. 4. Let $f_{2_{\text{in}}}$ and $f_{2_{\text{out}}}$ be the tangential forces applied to gear 2 from gears 1 and 3, respectively, and; $f_{\text{gb}}$ be the force applied to gear 2 from the gearbox; $\boldsymbol{f}_2 = [f_{2_{\text{in}}}, f_{2_{\text{out}}}, f_{\text{gb}}]^T$. Then, the moment balance in gear 2 is expressed as follows:

$$\begin{bmatrix} l_{2a} & l_{2b} & -l_{2c} \\ r_{2_{\text{in}}} & -r_{2_{\text{out}}} & 0 \end{bmatrix} \boldsymbol{f}_2 = \boldsymbol{0} \quad (1)$$

where $l_{2a}, l_{2b},$ and $l_{2c}$ denote the lengths from the rotational axis of gear 1 to the acting points of $f_{2_{\text{in}}}, f_{2_{\text{out}}},$ and $f_{\text{gb}}$, respectively, and $r_{2_{\text{in}}}$ and $r_{2_{\text{out}}}$ denote the pitch radii of the input gear meshing with gear 1 and of the output gear meshing with gear 3 in double gear 2, respectively (Fig. 4(a)). Next, the moment balance at gear 4 is considered. The tangential force applied to gear 3 by gear 4 is denoted as $f_{2_{\text{out}}}$ (Fig. 4(b)). If $\tau_4$ is the torque applied to gear 4 about its rotational axis by gear 3, $\tau_4 = r_4 f_{2_{\text{out}}}$, where $r_4$ is the pitch radius of gear 4. When force $f_{tip}$ is applied to the fingertip by the object, the torque applied to gear 4 by $f_{tip}$ is balanced with $\tau_4$ as follows:

$$\tau_4 = r_4 f_{2_{\text{out}}} = l_{tip} f_{tip} \quad (2)$$

where $l_{tip}$ denotes the moment arm from the rotational axis of gear 4 to the acting point of $f_{tip}$. Finally, from the moment balance at gear 1, the input motor torque ($\tau_m$) is given as

$$\tau_m = l_{2b} f_{2_{\text{in}}} \quad (3)$$

From (1) and (2), the relationship between $\tau_m$ and $f_{tip}$ is

$$\tau_m = \alpha_1 f_{tip} \quad (4)$$
$$\alpha_1 = (l_{tip} l_{2b} r_{2_{\text{out}}})/(r_{2_{\text{in}}} r_4) \quad (5)$$

Torque $\tau_{\text{gb}}$ applied to the preloaded gearbox determines the torque needed to twist the rotation unit. From (1) and (2), $f_{\text{gb}}$ can be expressed as follows:

$$f_{\text{gb}} = \alpha_2 f_{tip} \quad (6)$$
$$\alpha_2 = \frac{l_{2a} r_{2_{\text{out}}} + l_{2b} r_{2_{\text{in}}}}{l_{2c} r_{2_{\text{in}}}} \frac{l_{tip}}{r_4} \quad (7)$$

Considering that the pair of double gears 2 is installed antipodally to the rotational axis of the gearbox, from (6), $\tau_{\text{gb}}$ is given as

$$\tau_{\text{gb}} = 2l_{c2} f_{\text{gb}} = 2l_{c2} \alpha_2 f_{tip} \quad (8)$$

If $\tau_{pl}^{max.sf}$ is the maximum static frictional torque applied to the gearbox by the preloader about the rotational axis of the gearbox (rotation unit), the condition for activating the twisting of the rotational unit is as follows (see Fig. 4(c)):

$$\tau_{\text{gb}} = 2l_{c2} \alpha_2 f_{tip} > \tau_{pl}^{max.sf} \quad (9)$$

From (4), the corresponding motor torque is given as

$$\tau_m = \frac{\alpha_1}{2l_{c2} \alpha_2} \tau_{pl}^{max.sf} = \tau_{th} \quad (10)$$

According to (9) and (10), the twisting motion of the gripper is generated when the tip force ($f_{tip}$) exceeds the $\tau_{pl}^{max.sf}/2l_{c2}\alpha_2$ value, which is determined by preload value $\tau_{pl}^{max.sf}$, as $\tau_m$ increases. From (9), the point at which the grasping mode is activated can be determined by tuning $\tau_{pl}^{max.sf}$.

After switching from grasping to twisting, the kinetic frictional torque $\tau_{pl}^{kf}$ from the preload is applied to the gearbox (rotation unit), and $\tau_{\text{gb}}$ becomes

$$\tau_{\text{gb}} = \tau_{pl}^{kf} \Rightarrow f_{tip} = \tau_{pl}^{kf}/(2l_{c2}\alpha_2) \quad (11)$$

From (4), the corresponding motor torque is

$$\tau_m = (\alpha_1 \tau_{pl}^{kf})/(2l_{c2}\alpha_2) = \tau_{const} \quad (12)$$

From (12), the motor torque remains constant after the switch to twisting unless an external load is applied. Hence, the gripper grasping mode (i.e., grasping or twisting motion) can be detected by monitoring the motor torque. An experiment was conducted to validate this analysis. During the experiment, the state of the gripper was observed when the motor operated at a constant angular velocity. The motor current was measured during the motor operation, and the motor torque was derived from the motor current, as described in [27]. The initial motor angle at which the gripper was fully opened was 0°. The results of the motor torque measurements and observed states of the gripper are shown in Fig. 5. The magnitude of the preload was set to "medium" (see Section II.F for the definition of "medium" and details regarding the effect of the preload value). The gripper was classified into three states based on motor torque measurements during the gripper operation. First, the fingers moved until the fingertips came into contact with each other. This state was referred to as "state A." In this state, the motor torque ($\tau_m$) approached zero, and the grasping motion was activated. Once the fingertips were in contact with each other, $\tau_m$ increased as the tip force increased. This state was referred to as "state B." In this state, neither the finger motion nor the twisting motion was activated. When $\tau_m$ reached $\tau_{th}$, the twisting motion was activated. At this point, $\tau_m$ became constant ($\tau_{const}$). This state was referred to as "state C." These results validated the aforementioned analysis.

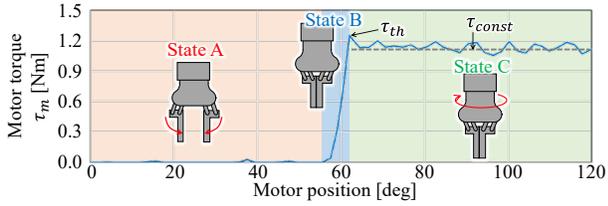

**Fig. 5.** Measured torque and observed states of the gripper at a constant motor speed

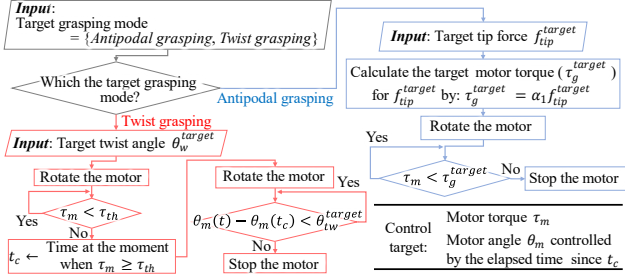

**Fig. 6.** Methodology for controlling the proposed gripper

### E. Methodology for Controlling Gripper

The proposed gripper requires two types of control methodologies: 1) motion control of the two fingers to grasp an object via antipodal grasping and 2) control of the rotational angle of the twisting motion. This section presents the methodologies based on the monitoring of the motor torque, as summarized in Fig. 6.

*1) Methodology for grasping without twisting*

As described in the previous subsection, the motor torque ($\tau_m$) increases until the twisting motion is activated. If the motor rotation stops before $\tau_m$ reaches the threshold value $\tau_{th}$ (see (10)), grasping can be achieved without twisting. The desired tip force ($f_{tip}^{target}$) is set such that $\alpha_1 f_{tip}^{target} = \tau_g^{target} < \tau_{th}$ (see (4)). Then, by adjusting $\tau_m$ to be $\tau_g^{target} = \alpha_1 f_{tip}^{target}$, the tip force of antipodal grasping can be adjusted to $f_{tip}^{target}$. The experimental validation to confirm whether the methodology can realize grasping without twisting under different preload values is presented in Section II.F.

*2) Methodology for controlling twist angle*

Because the payload due to twist grasping is determined by the extent of wrapping (see Section III for details), the twist angle must be controlled. The twist angle must also be controlled in other operations, such as the alignment of the object posture. In this subsection, the control method is presented. The motor is assumed to operate at a constant angular velocity. Let $t_c$ be the time when the motor torque ($\tau_m$) exceeds $\tau_{th}$ after grasping, i.e., the time at which the twisting motion begins (see Fig. 5). The time elapsed since $t_c$ corresponds to the rotational angle of the wrist. Let $\theta_m(t)$ be the motor angle at time $t$, $\theta_{tw}^{target}$ be the desired twist angle, and $t_d$ be $t$ such that $\theta_{tw}^{target} = \theta_m(t) - \theta_m(t_c)$. By monitoring the motor torque to detect $t_c$ and stopping the motor at $t_d$, $\theta_{tw}^{target}$ can be obtained (Fig. 6). Note that $\tau_{th}$ is determined by the preload value, as shown in (10), and $\theta_{tw}^{target}$ can be set to an unlimited positive value. This methodology was used to realize twist grasping for a large payload with the desired number of wraps. We conducted experiments to evaluate the posturing accuracy of twisting motion using this methodology. The experimental setup is shown in Fig. 7. The rotational displacement of the object posture ($\Delta\theta_{obj}$) after twisting is given as follows:

$$\Delta\theta_{obj} = \theta_{obj}^{after} - \theta_{obj}^{before} \quad (13)$$

where $\theta_{obj}^{after}$ and $\theta_{obj}^{before}$ denote the postures (angles) of the object before and after the twisting, respectively. The camera was placed below the object under a clear acrylic plate. The object postures about the $z_A$ axis were derived by processing the images captured by the camera (Fig. 7). We conducted the experiments three times for each condition, with $\theta_{tw}^{target}$ values of 90°, 180°, 270°, and 360°. The preload value was set to "medium," as shown in Fig. 5, and the threshold $\tau_{th}$ was set to 1.1 Nm (see Section II.F for details). The results are summarized in Table I, and representative profiles of the motor torque $\tau_m$ and motor angle $\theta_m$ during finger closing, grasping, and twisting are shown in Fig. 8. The threshold $\tau_{th}$ was lower than the actual $\tau_{th}$ because the threshold was determined such that the transition to state C could always be detected. The preload value represented by $\tau_{pl}^{max.sf}$ in (10) and $\tau_{pl}^{kf}$ in (12) corresponds to $\tau_m$ in (10) and (12), respectively. Either $\tau_{pl}^{max.sf}$ or $\tau_{pl}^{kf}$ can be used to tune the preload value, because they

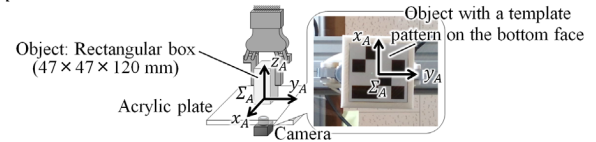

**Fig. 7.** Experimental setup to evaluate the posturing accuracy during twisting

### TABLE I
### RESULTS FOR THE POSTURING ACCURACY IN TWISTING

|  | Target rotational angle of the gripper $\theta_{tw}^{target}$ [deg] | | | |
| --- | --- | --- | --- | --- |
|  | 90 | 180 | 270 | 360 |
| $\Delta\theta_{obj}$ [deg] | 89.5 | 179.6 | 269.9 | 359.7 |
| Standard deviation | 0.2 | 0.3 | 0.6 | 0.3 |

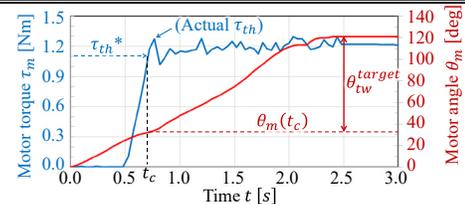

**Fig. 8.** Representative results of the proposed methodology ($\theta_{tw}^{target}$ was set to 90°) for the motor torque $\tau_m$ and motor angle $\theta_m$ during finger closing, grasping, and twisting

\* This value is lower than the actual $\tau_{th}$ value because the threshold was determined such that the transition to state C could always be detected.

correspond to each other. If $\tau_{pl}^{kf}$ is used to set the preload value, the preload value can be set by tuning the motor torque during the twisting motion to be equal to the motor torque derived by substituting the desired preload value into (12). The use of $\tau_{pl}^{kf}$ allows tuning with a motor torque that is measured as almost a constant value for a while, whereas the method based on $\tau_{pl}^{max.sf}$ uses a torque that is measured only momentarily. Therefore, the method using $\tau_{pl}^{kf}$ can tune the preload value (and the corresponding $\tau_{th}$) more accurately (with a smaller error) than that using $\tau_{pl}^{max.sf}$. In addition, the tuning is easier. Hence, the method using $\tau_{pl}^{kf}$ was adopted. The actual point where the twisting motion was activated differed from that obtained via the method using $\tau_{pl}^{kf}$, as shown in Fig. 8. However, the difference was small and had little effect on the posturing accuracy. The results presented in Table I demonstrate that the proposed methodology yields a high posturing accuracy during twisting.

*F. Effect of Preload Value*

This section describes the effects of the preload value on the tip force and motor torque during gripper twisting. First, the tip forces at different preload values were investigated. The experimental setup is shown in Fig. 9(a). The gripper was operated using the methodology shown in Fig. 6 to perform antipodal grasping. The fixed force gauge measured the tip force when finger motion was stopped using the proposed methodology. In the experiments, the preload values were set to three levels: low, medium, and high. The levels were determined to be within the range in which the desired behavior (switching from grasping to twisting motion) was stably achieved via trial and error. The motor torque $\tau_g^{target}$ was set between 0.04 and 1.60 Nm in 0.17-Nm intervals for each preload value. The experiments were conducted three times for each condition. For comparison, the tip force was also theoretically derived from (4) by setting $\tau_m$ to each target motor torque ($\tau_g^{target}$). The parameters were set as follows: $l_{tip}$ = 50, $l_{2b}$ = 18, $r_{2\,out}$ = 10, $r_{2\,in}$ = 12, and $r_4$ = 10 (mm). The results are shown in Fig. 9(b). The measured tip-force values were approximately identical to the theoretical values. When $\tau_g^{target}$ was set to a larger value, a larger magnitude of $f_{tip}$ was obtained. If $\tau_g^{target}$ was greater than $\tau_{th}$, a twisting motion is generated before the motor stops. Hence, the tip force for antipodal grasping should be controlled within a range where $\tau_g^{target} < \tau_{th}$, as determined by the preload value. The threshold $\tau_{th}$ value obtained from the results in Fig. 9(b) is presented in Table II for each preload value. Next, we considered the torque $\tau_{tw}$ that could be applied externally during the twisting motion. $\tau_{tw}$ is required, for example, during screw tightening, and is generated in addition to $\tau_{const}$ during twisting. The torque was derived by measuring the pushing force applied at a position 44 mm from the center of the rotation axis of the rotation unit, as shown in Fig. 10(a). As illustrated in Fig. 10(b), the experimental procedure was as follows: 1) the initial posture of the gripper was set such that the posture could be rotated by 90° from the initial posture to push the force gauge; 2) the motor was rotated at a constant velocity until the gripper was stopped by contact with the force gauge (the state of the gripper shifted from state A to B and then to C, and twisting was activated in state C); and 3) the motor was stopped when the motor torque reached its maximum value. The maximum torque (denoted as $\tau_{tw}$) was derived using the force value of the force gauge when the gripper was stopped, i.e., $\tau_{tw}$ was calculated as the product of the force value and the moment arm (44 mm). The experiments were conducted for each of the three preload levels. The results are shown in Fig. 11. As shown in Fig. 11(a), $\tau_{tw}$ increased as the preload value was decreased. The joint twisting torque ($\tau_{const}$) was derived by averaging the motor torque in state C, as shown in Fig. 11(b). Fig. 11(a) shows a stacked bar chart of $\tau_{const}$ and $\tau_{tw}$. The results indicated that a larger preload induced a larger $\tau_{const}$. The sum of $\tau_{tw}$ and $\tau_{const}$ was approximately equal to the maximum motor torque ($\tau_m^{max}$) under all the preload conditions. The magnitude of $\tau_{const}$ indicated the magnitude of the generable tip force (see (11) and (12)). The results indicated a tradeoff between the additional torque during twisting and the additional tip force. The preload value must be tuned according to the target application.

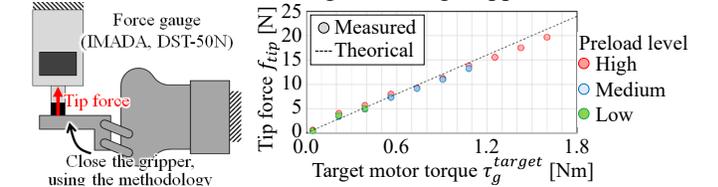

(a) Experimental setup    (b) Results

**Fig. 9.** Evaluation of the tip force with changes in the preload level. The tip-force values before the activation of the twisting motion are plotted against the target motor torque.

TABLE II
$\tau_{th}$ FOR SWITCHING GRIPPER MODE AT THREE PRELOAD LEVELS

|  | Preload level | | |
|---|---|---|---|
|  | Low | Low | Low |
| $\tau_{th}$ [Nm] | 0.4 | 0.4 | 0.4 |

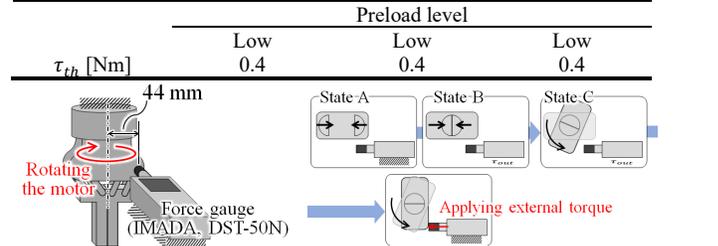

(a) Experimental setup    (b) Procedure (bottom view)

**Fig. 10.** Experimental setup and procedure for measuring the additional torque during twisting

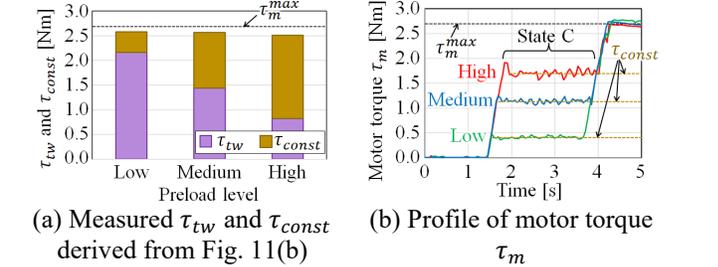

(a) Measured $\tau_{tw}$ and $\tau_{const}$    (b) Profile of motor torque
derived from Fig. 11(b)                         $\tau_m$

**Fig. 11.** Relationship between the additional generable torque during twisting ($\tau_{tw}$) and the torque from the twisting motion ($\tau_{const}$) and the corresponding profile of the motor torque $\tau_m$

## III. Feasibility Analysis of Twist Grasping

Twist grasping is a special grasping mode provided by the proposed gripper. In this section, an analysis of the payload in twist grasping is presented and experimentally validated. The procedure for grasping flexible thin objects via twist grasping is illustrated in Fig. 1(b). The extent to which the object is wrapped is controlled using the methodology shown in Fig. 6.

### A. Feasibility Analysis

A model for the case in which the gripper holds an object via twist grasping with $n$ wraps is illustrated in Fig. 12. The base coordinate frame ($\Sigma_B$) for the analysis was set, as shown in Fig. 12. In this analysis, we considered the case in which the number of wraps ($n$) satisfied $n \in \{\mathbb{N}, 2 \leq n\}$. The analysis focused on a small section with a width of $d\theta$ located at $\theta$, which is the angle from the starting point of the object's wrapped section, with a positive direction corresponding to the direction of the $+x_B$ axis (see Fig. 12). The wrapped object was separated into two sections: section A (blue section in Fig. 12) and section B (green section). Both the inside and outside of section A were in contact with the other sections (i.e., section B, the fingertip, and other parts of section A), with a range of $0 \leq \theta < 2(n-2)\pi + \frac{3}{2}\pi$. In contrast, only the interior of section B was in contact with section A. The range of section B was $2(n-2)\pi + \frac{3}{2}\pi \leq \theta < 2(n-1)\pi + \frac{3}{2}\pi$. Let $f_t$ and $f_t + df_t$ be the tensions applied to the end points of $\theta$ and $\theta + d\theta$ in the small section, respectively, and $f_{os}$ and $f_{is}$ be the forces applied to the small section by the other sections in contact with its outside and inside surfaces, respectively. The friction coefficients between the object surface and the fingertip and between the object surfaces in contact were assumed to be equal. The force balances in the radial and tangential directions in sections A and B are expressed as follows:

Section A:
$$f_{is} = f_{os} + (f_t + df_t)\sin\frac{d\theta}{2} + f_t\sin\frac{d\theta}{2}$$
$$(f_t + df_t)\cos\frac{d\theta}{2} + \mu f_{os} = f_t\cos\frac{d\theta}{2} + \mu f_{is} \quad (14)$$

Section B:
$$f_{is} = (f_t + df_t)\sin\frac{d\theta}{2} + f_t\sin\frac{d\theta}{2}$$
$$(f_t + df_t)\cos\frac{d\theta}{2} = f_t\cos\frac{d\theta}{2} + \mu f_{is}$$

If $f_g$ is the force acting at the starting point of the wrapping and corresponds to the payload due to antipodal grasping, $f_b$ is the tension applied to the boundary area between sections A and B, and $f_{obj}$ is the payload, the following relationships are obtained from (14):

$$\mu \int_0^{2(n-2)\pi + \frac{3}{2}\pi} d\theta = \int_{f_g}^{f_b} \frac{df_t}{f_t} \Rightarrow f_b = e^{2\mu\left(n - \frac{5}{4}\right)\pi} f_g \quad (15)$$

$$\mu \int_{2(n-2)\pi + \frac{3}{2}\pi}^{2(n-1)\pi + \frac{3}{2}\pi} d\theta = \int_{f_b}^{f_{obj}} \frac{df_t}{f_t} \Rightarrow f_{obj} = e^{2\mu\pi} f_b \quad (16)$$

From (15) and (16), the payload from $n$ rotations of the gripper is given as

$$f_{obj} = \beta_1 \beta_2^n f_g \quad (\beta_1 = e^{-\frac{1}{2}\mu\pi}, \beta_2 = e^{2\mu\pi}) \quad (17)$$

Note that (17) represents the case in which $n \geq 2$. If the number of wraps is $< 2$, the payload can be obtained in the same aforementioned manner by changing the range of θ in sections A and B. According to (17), the payload increases with the number of wraps, and the gripper can grasp heavy objects via twist grasping as long as it does not break.

### B. Validation

In this section, the theoretical payload due to twist grasping using various numbers of wraps is validated. First, we evaluated the payload without twisting, i.e., for conventional antipodal grasping, to determine $f_g$, which was needed to derive the payload ($f_{obj}$) due to twist grasping (see (17)). The experimental setup is shown in Fig. 13. A flexible thin object (packing tape) was grasped via antipodal grasping using the methodology shown in Fig. 6. The gripper was moved upward using an automatic positioning stage. The payload ($f_g$) was defined as the magnitude of the force when slippage occurred between the fingertip and the object and was measured using a force gauge. The experiments were conducted three times for each preload level (low, medium, and high). The target motor torque ($\tau_g^{target}$) was set to the threshold value ($\tau_{th}$), which provided the maximum tip force for each preload level, to switch the grasping mode (Table II). The results are presented in Table III. Next, the payloads due to twist grasping with $n$ wraps were derived theoretically using the obtained values of $f_g$ and (17). The friction coefficient ($\mu$) was assumed to be 0.3. The results for $n = 0.5, 1, 2,$ and 3 are summarized in Table IV. Finally, we experimentally verified that the proposed gripper could hold the derived payloads via twist grasping. The experimental setup is shown in Fig. 14. The target object was packing tape with a variable weight attached to the end. The value of the weight was set such that the weight of the target object was between the values listed in Table IV. The methodology shown in Fig. 6 was adopted with the threshold values ($\tau_{th}$) listed in Table II to achieve the desired number of

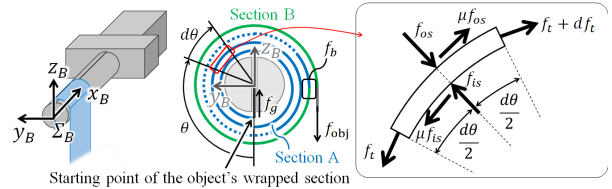

**Fig. 12.** Model of the object grasped via twist grasping

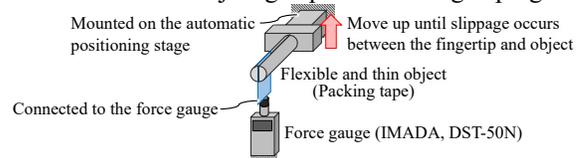

**Fig. 13.** Experimental setup for measuring the payload without twisting

### TABLE III
### Payload Without Twisting $f_g$

|  |  | Preload level | | |
|---|---|---|---|---|
|  |  | Low | Medium | High |
| $f_g$ | Mean [N](kgf) | 2.2 (0.2) | 3.9 (0.4) | 5.7 (0.6) |
|  | Standard deviation | 0.3 | 0.5 | 0.4 |

TABLE IV
ESTIMATED PAYLOAD OWING TO TWIST GRASPING ($f_{obj}$)

| Num. of wraps $n$ | Amplification ratio ($\beta_1 \beta_2^n$)* | $f_{obj}$ [N](kgf) Low preload | Medium preload | High preload |
|---|---|---|---|---|
| 0.5 | 1.6 | **3.5 (0.4)** | **6.3 (0.6)** | **9.2 (0.9)** |
| 1 | 4.1 | **9.0 (0.9)** | **16.0 (1.6)** | **23.6 (2.4)** |
| 2 | 27.1 | **59.6 (6.1)** | **105 (10.8)** | 155 (15.8) |
| 3 | 178 | 392 (40) | 695 (71) | 1022 (104) |

\* The friction coefficient is assumed to be 0.3.
**Bold values**: Validated using actual experiments

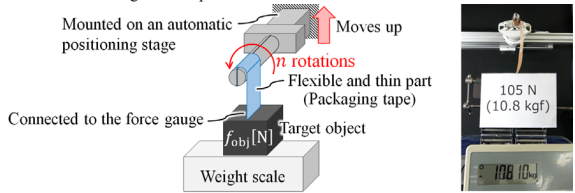

**Fig. 14.** Experimental setup to determine the payload for twist grasping

wraps ($n$). After the gripper achieved the target posture, it lifted the object using an automatic positioning stage. Considering the durability of the gripper and stage, the experiments were conducted with a target object weighing <11 kg (see the bold values in Table IV). Under all the preload conditions, the object was lifted (see the attached video). The results validated the proposed model and confirmed that the proposed gripper can grasp a heavy object (>10 kg) via twist grasping.

## IV. EXPERIMENTAL VALIDATION

The proposed gripper was experimentally evaluated. In the experiments, the preload level was set to "medium." Correspondingly, the threshold $\tau_{th}$ (Table II) was set as 1.1 Nm to control the gripper.

### A. Grasping Test

The proposed gripper was evaluated via grasping tests with antipodal and twist grasping.

*1) Antipodal grasping*

The proposed gripper, which was attached to an automatic positioning stage, grasped an object (Fig. 15(a)) placed on a table using the methodology shown in Fig. 6. Various objects, including a heavy bottle (0.6 kg), spherical baseball, small binder clip (width of 18 mm and height of 9 mm), and deformable toy doll, were successfully grasped, as shown in Fig. 15(b). Moreover, owing to the preload, the gripper mechanism allowed the finger to move before activating the twisting motion of the wrist. Hence, a release motion without twisting was achieved by restoring the motor to its initial state (Fig. 15(c)).

*2) Twist grasping*

The experimental setup is shown in Fig. 16(a). The target objects were a laundry-detergent laminated pouch (1.6 kg) and a rice pouch (10 kg). Rice pouches are among the heaviest pouches in Japanese supermarkets. The proposed gripper could grasp these target objects from a table by using the methodology shown in Fig. 6. According to the results presented in Table IV, the pouch with detergent was grasped in one rotation of the twist grasping mechanism, and the pouch with rice was grasped in two rotations. The external load caused by the abrupt lifting of the object during twisting must be minimized to obtain the desired number of wraps. Hence, the gripper body (positioning stage) moved downward during twisting (see Fig. 16(a)). If $r_f$ is the radius of the finger, the displacement speed $v$ of the positioning stage during twisting can be calculated as

$$v = r_f \dot{\theta}_m \quad (18)$$

Using this strategy, both objects were successfully grasped (Figs. 1(b) and 16(b)). During release, the objects should be unrolled to open the gripper. This was accomplished by rotating the motor in the reverse direction because reverse twisting was generated before the fingers opened; the fingers could not open when the object was wrapped around them (Fig. 17).

In summary, both antipodal grasping and twist grasping were achieved using the proposed methodology, as shown in Fig. 6, and a release mechanism was achieved by restoring the motor to its initial state.

### B. Operation Test

Several operations were conducted using the proposed gripper to evaluate the efficiency of the wrist twisting mechanism. The tests included 1) tightening and untightening a bolt using a special tool (Fig. 18(a)) and 2) winding a sheet of paper using two manipulators equipped with the proposed grippers, twisting in opposite directions. The other tests are shown in the video clip. Both operations were successfully performed. The tightening and untightening of a bolt that requires twisting in bilateral directions was achieved using a

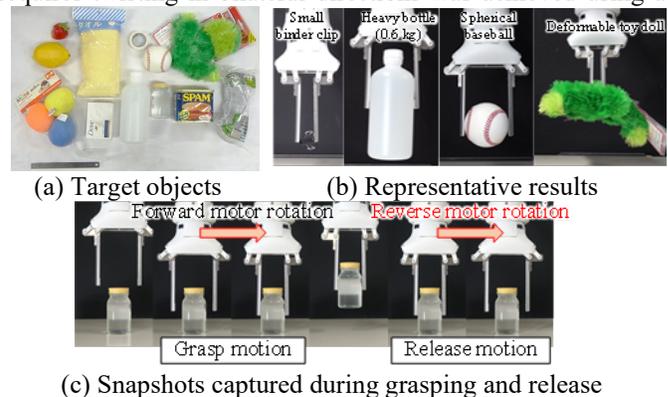

(a) Target objects    (b) Representative results

(c) Snapshots captured during grasping and release

**Fig. 15.** Target objects and the antipodal grasping test results

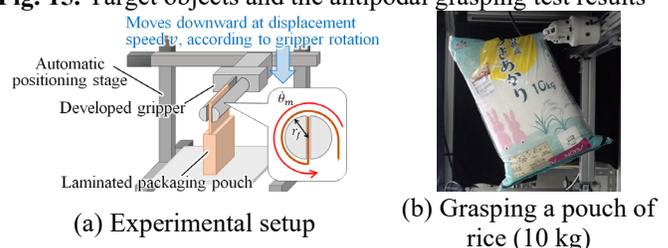

(a) Experimental setup    (b) Grasping a pouch of rice (10 kg)

**Fig. 16.** Experimental setup and results for grasping a laminated pouch of rice (10 kg) via twist grasping

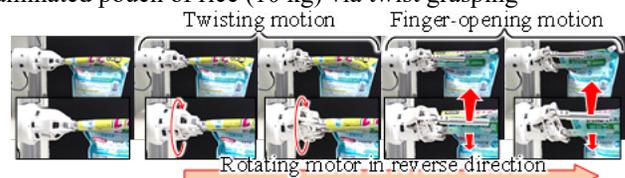

**Fig. 17.** Snapshots during the release process of twist grasping

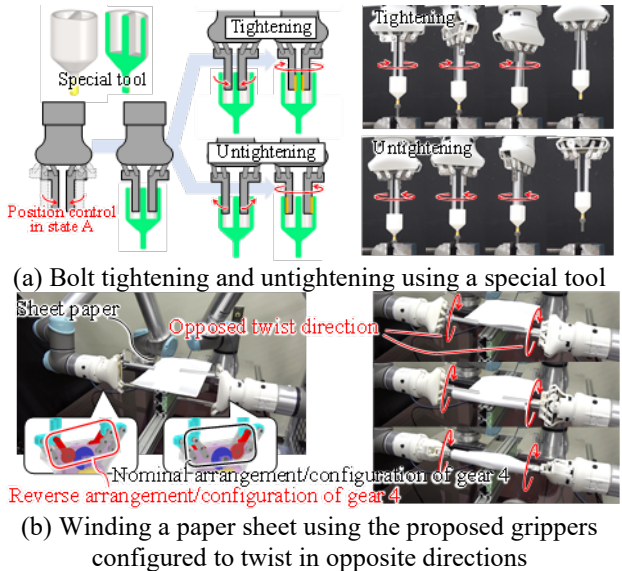

(a) Bolt tightening and untightening using a special tool

(b) Winding a paper sheet using the proposed grippers configured to twist in opposite directions

**Fig. 18.** Representative results of the operation tests

special tool (Fig. 18(a)). In the second operation, to reverse the twisting direction of the gripper installed on one of the manipulators, the configuration of gear 4 was reversed relative to the nominal configuration, as shown in Fig. 3. With this modification, the two manipulators successfully completed the operation (Fig. 18(b)).

## V. CONCLUSION

This paper presents a novel robotic gripper that realizes the grasping and twisting motions of the wrist using a single actuator. Pertinent control methodologies are also presented. A lightweight gripper (0.4 kg) was designed. The gripper achieved a special grasping mode (twist grasping) that involved wrapping a flexible thin object around the fingers by twisting the gripper. The analysis and experimental validation of the gripper indicated that twist grasping provided a large payload that increased with the number of wraps. It was demonstrated that the proposed gripper prototype could grasp a laminated pouch weighing 10 kg via twist grasping. Several grasping and operation tests were conducted to verify the effectiveness and usefulness of the proposed gripper. The proposed control methodologies for grasping and twisting were validated by performing grasping and operation tests. A release mechanism for the gripper was also realized. Although the twist direction of the gripper was limited to the set direction, the operation tests revealed that this limitation can be overcome by using a special tool and collaboratively using grippers with opposite twist directions. The switching points of the different grasping modes were manually tuned using a preloader. In the future, we plan to incorporate a mechanism for adjusting the preload level without additional actuators, for example, by relying on contact with the environment. The issue of sliding wear due to friction between the gearbox and the indenter can reduce the preload value. In the proposed design, friction occurs between the aluminum plate installed in the gearbox and the resin indenter. The indenter has been designed to be easily replaceable because it wears more easily than the gearbox. Our future work will also involve countermeasures such as increasing the contact area of the indenter to reduce the degree of wear caused by the preload.